\documentclass[sigconf]{acmart}

\AtBeginDocument{%
  \providecommand\BibTeX{{%
    \normalfont B\kern-0.5em{\scshape i\kern-0.25em b}\kern-0.8em\TeX}}}

\setcopyright{acmcopyright}
\copyrightyear{2018}
\acmYear{2018}
\acmDOI{10.1145/1122445.1122456}

\acmConference[SSL@WWW'21]{SSL@WWW'21: The Workshop on Self-Supervised Learning for the Web}{April 19--23, 2021}{Virtual Conference}
\acmBooktitle{SSL@WWW'21: The Workshop on Self-Supervised Learning for the Web,
  April 19--23, 2021, Virtual Conference}
\acmPrice{15.00}
\acmISBN{978-1-4503-XXXX-X/18/06}

\usepackage{amsmath, amsfonts,amsthm}
\usepackage{xspace, color, latexsym}
\usepackage{subfig, microtype, stmaryrd, tikz, tikz-qtree, booktabs}
\usepackage{mathtools}
\usepackage{multirow}
\usepackage[noend,ruled,linesnumbered]{algorithm2e} 

\newcommand{\pair}[1]{\langle #1 \rangle}

\def \C {\mathcal{C}}

\def \E {\mathcal{E}}

\def \N {\mathcal{N}}
\def \O {\mathcal{O}}

\def \T {\mathcal{T}}

\begin{document}

\newcommand{\TaxoExpan}{\mbox{\sf TaxoExpan}\xspace}
\newcommand{\Arborist}{\mbox{\sf Arborist}\xspace}
\newcommand{\TaxoOrder}{\mbox{\sf TaxoOrder}\xspace}
\newcommand{\TaxoOrderBf}{\mbox{\sf \textbf{TaxoOrder}}\xspace}
\newcommand{\Name}[1]{\mbox{\sf #1}\xspace}
\newcommand{\NameBf}[1]{\mbox{\sf \textbf{#1}}\xspace}
\newtheorem{thm:def}{Definition}

\title{Who Should Go First? A Self-Supervised Concept Sorting Model for Improving Taxonomy Expansion}


\author{Xiangchen Song}
\affiliation{%
  \institution{Department of Computer Science\\University of Illinois at Urbana-Champaign}
  \streetaddress{}
  \city{Champaign}
  \state{IL}
  \country{USA}}
\email{xs22@illinois.edu}
\author{Jiaming Shen}
\affiliation{%
  \institution{Department of Computer Science\\University of Illinois at Urbana-Champaign}
  \streetaddress{}
  \city{Champaign}
  \state{IL}
  \country{USA}}
\email{js2@illinois.edu}
\author{Jieyu Zhang}
\affiliation{%
  \institution{Paul G. Allen School of Computer Science \& Engineering\\University of Washington}
  \streetaddress{}
  \city{Seattle}
  \state{WA}
  \country{USA}}
\email{jieyuz2@cs.washington.edu}
\author{Jiawei Han}
\affiliation{%
  \institution{Department of Computer Science\\University of Illinois at Urbana-Champaign}
  \streetaddress{}
  \city{Champaign}
  \state{IL}
  \country{USA}}
\email{hanj@illinois.edu}
\renewcommand{\shortauthors}{Song, et al.}

\begin{abstract}
  Taxonomies have been widely used in various machine learning and text mining systems to organize knowledge for facilitating downstream tasks. 
One critical challenge is that, as data and business scope grow in real applications, existing taxonomies need to be expanded to incorporate new concepts.
Previous work on taxonomy expansion assumes those concepts are independent and process them one after another.
As a result, they ignore the potential relationships among new concepts.
However, in reality, those new concepts tend to be correlated and form hypernym-hyponym structures.
In such a scenario,  ignoring the relations among new concepts and inserting them into the taxonomy in an arbitrary order may trigger error propagation.
For example, previous taxonomy expansion systems may insert hyponyms to existing taxonomies before their hypernyms, leading to sub-optimal expanded taxonomies.
To complement existing taxonomy expansion systems, we propose \TaxoOrder, a novel self-supervised framework that simultaneously discovers hypernym-hyponym relations among new concepts and decides their insertion order.
\TaxoOrder can be directly plugged into any taxonomy expansion system and improve the quality of expanded taxonomies. 
Experiments on two real-world datasets validate the effectiveness of \TaxoOrder to enhance taxonomy expansion systems, leading to better-resulting taxonomies with comparison to baselines under various evaluation metrics.

\end{abstract}


\maketitle

\DeclarePairedDelimiter\ceil{\lceil}{\rceil}
\begin{figure}[ht]
    \centering
    \includegraphics[width=\linewidth]{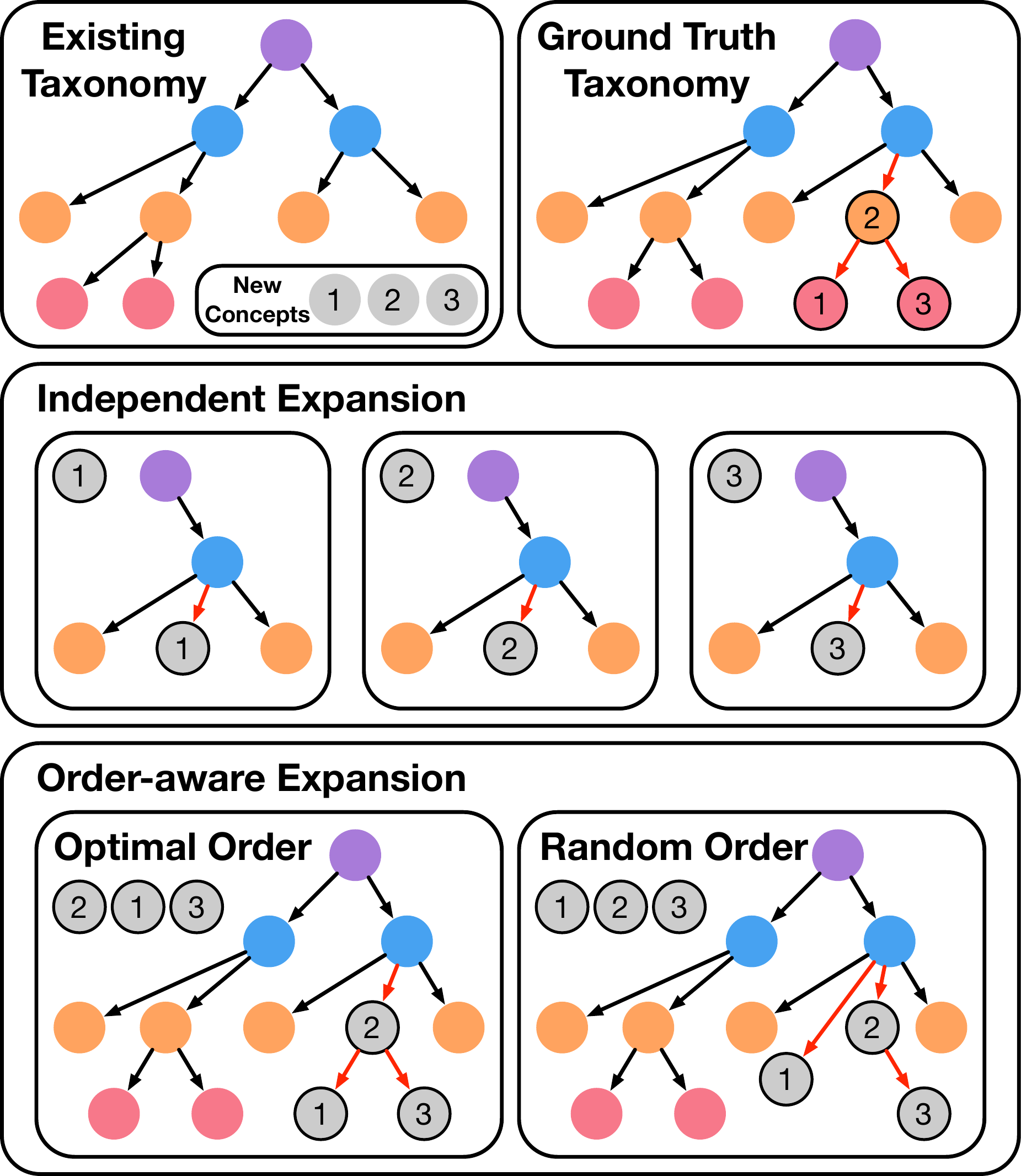}
    \caption{An illustrative example of the taxonomy expansion task. \textbf{(Top-left)}: an existing taxonomy with three new concepts. (Top-right) the ground truth taxonomy we want to discover. \textbf{(Middle)}: independent expansion, where the taxonomy expansion task is divided into 3 independent tasks. Each new concepts can only be inserted as children of existing taxonomy concepts. No hypernym-hyponym relation among new concepts can be discovered. \textbf{(Bottom)}: order-aware expansion, we illustrate the expanded taxonomy with optimal and random order of concepts to be inserted in order-aware taxonomy expansion.}
    \label{fig:task}
\end{figure}

\section{Introduction}
Taxonomies, represented as Direct Acyclic Graphs (DAGs), have been used to organize knowledge and information for centuries \cite{stewart2008building}.
High-quality taxonomies can benefit many downstream applications such as query understanding \cite{hua2016understand, yang2020taxogan}, content browsing \cite{yang-2012-constructing}, personalized recommendation \cite{zhang2014taxonomy,Huang2019TaxonomyAwareMR}, and user-behavior modeling \cite{Menon2011ResponsePU}.
In the past, the majority of taxonomies are manually curated by human experts. 
Such curation is time-consuming and labor-intensive.
To reduce the burden of human experts, many automatic taxonomy construction methods \cite{Mao2018EndtoEndRL,Shen2018HiExpanTT,Zhang2018TaxoGenUT} have been proposed.
With the growth of human knowledge comes the increasing demand for expanding the existing taxonomies to incorporate new concepts.
Driven by this demand, previous studies \cite{Zhang2021TaxonomyCV,Yu2020STEAMST,Mao2020OctetOC,Shen2020TaxoExpanST,Manzoor2020ExpandingTW} on taxonomy expansion aim to rank the candidate hypernyms in existing taxonomies and insert new concepts as hyponyms of the most likely hypernym.

In reality, the order of inserting operations for new concepts is critical for existing taxonomy expansion systems.
Suppose that a hyponym concept is inserted before its hypernyms, existing taxonomy expansion systems can hardly recover the ground truth hypernym-hyponym structure, because later concepts can only be added as leaf nodes in taxonomy.
For example, considering a hypernym-hyponym pair of concepts (``geometry'', ``rectangle''), if ``rectangle'' is first inserted into the existing taxonomy, then when processing ``geometry'', we can only insert it as a hyponym of ``rectangle'', which is incorrect. 
Figure \ref{fig:task} (Bottom) illustrates how inserting order determines the quality of expanded taxonomy. 
For an optimal taxonomy expansion model which always outputs the ``correct'' hypernym concept, if the inserting order is sub-optimal (1-3-2 or 3-1-2), it can never recover ground truth taxonomy.
To bypass such a problem, existing taxonomy expansion systems focus on a simplified task: they divide the task of inserting k new concepts into k \emph{independent} tasks whose input set of new concepts contains only one element.
As illustrated in the ``Independent Insertion" case in Figure~\ref{fig:task} (Middle), this strategy ignores the potential dependencies of new concepts and restricts one new concept to be inserted underneath another new concept, which simplifies the taxonomy expansion task but passes the burden to human experts.

To address this issue, we instead study the \emph{order-aware} taxonomy expansion task as shown in Figure~\ref{fig:task} (Bottom).
We aim to discover the appropriate order of insertion and iteratively insert each individual new concept into the existing taxonomy.
We propose a novel sorting model \TaxoOrder, which determines an insertion order of new conecpts.
Within this order, hypernyms are ranked in front of hyponyms and thus when hyponyms are inserted, their potential parent nodes are already in the partially expanded taxonomy.
\TaxoOrder first learns the hypernym-hyponym relations from the existing taxonomy. Then it utilizes the learned model together with heuristic patterns to generate pseudo-edges of new concepts, which carries the relative hypernym-hyponym information between each concept pair. Finally, we decide the insertion order by topological sorting on the DAG constructed from pseudo-edges. \TaxoOrder can be integrated into any taxonomy expansion system for improving the performance of taxonomy expansion. 
On real-word academic concept taxonomies, we validate that \TaxoOrder can generate high-quality concept order and benefit taxonomy expansion models.

\section{Problem Formulation}
In this section, we first define a taxonomy and then formulate the ordered taxonomy expansion problem with an emphasis on the scope of this work.

\smallskip
\noindent \textbf{Taxonomy.}
A taxonomy $\mathcal{T=(N,E)}$ is a directed acyclic graph (DAG) where each node $n\in \mathcal{N}$ represents a concept and each directed edge $\langle n_p, n_c\rangle \in \mathcal{E}$ indicates a relation expressing that concept $n_{p}$ is the most specific concept that is more general than concept $n_{c}$. 
We refer to $n_{p}$ as the ``parent'' or ``hypernym'' of $n_{c}$ and $n_{c}$ as the ``child'' or ``hyponym'' of $n_{p}$.

\smallskip
\noindent \textbf{Taxonomy Expansion.}
In taxonomy expansion task, the input is (1) an existing taxonomy $\mathcal{T}^0=(\mathcal{N}^0,\mathcal{E}^0)$, and (2) a set of new concepts $\mathcal{C}$, either provided by users or automatically extracted from unstructured text. 
Our goal is to insert all new concepts into the existing taxonomy $\mathcal{T}^0$ and obtain an expanded taxonomy $\mathcal{T}^1 = (\mathcal{N}^0 \cup \mathcal{C}, \mathcal{E}^0 \cup \mathcal{R})$, where $\mathcal{R}$ is the set of hypernym-hyponym relations discovered by algorithm and each includes a new concept in $\mathcal{C}$ as hyponym. 

Previous studies \cite{Shen2020TaxoExpanST,Manzoor2020ExpandingTW,Yu2020STEAMST,Zhang2021TaxonomyCV} focus on a simplified version of the above problem: assume that the input set of new concepts contains only one element (i.e., $|\C|=1$), so that for any discovered hypernym-hyponym pair $\langle n_p, n_c \rangle \in \mathcal{R} $, we must have $n_p \in \mathcal{N}^0 $.
By this way, the original taxonomy expansion task is divided into $|C|$ independent simplified tasks.
In contrast, we focus on the original taxonomy expansion task in an iterative setting:
at iteration $t$, we have current taxonomy $\mathcal{T}^t = (\mathcal{N}^t, \mathcal{E}^t)$, then for the $t$-th concept $n^t$ in $\C$, we aim to find its most likely hypernym $n_p\in\N^t$ and the new taxonomy is $\mathcal{T}^{t+1} = (\mathcal{N}^{t+1}=\mathcal{N}^{t}\cup\{n^t\}, \mathcal{E}^{t+1}=\mathcal{E}^{t}\cup\{\langle n_p, n^t \rangle\})$.
As a result, an appropriate ordering of new concept is required.

\smallskip
\noindent \textbf{Concepts Sorting.}
The goal of concept sorting is to assign an appropriate order of new concepts in $\C$ for the iterative taxonomy expansion described above.
Intuitively, hypernyms are supposed to be inserted before the hyponyms, so that when processing hyponyms, their ground truth hypernyms are already in the current taxonomy. To this end, we aim to learn a concept sorting model which outputs the order of new concepts in $\C$ based on self-supervision generated from initial taxonomy $\T^0$.

\smallskip
\noindent \textbf{Scope of Study.}
We do not modify the taxonomic relations in the existing taxonomy, i.e., $\mathcal{E}$,  since the quality of the existing taxonomy is good enough and modifying existing taxonomy is less frequent and requires great efforts from human curators with caution.
\begin{figure*}
    \centering
    \includegraphics[width=\linewidth]{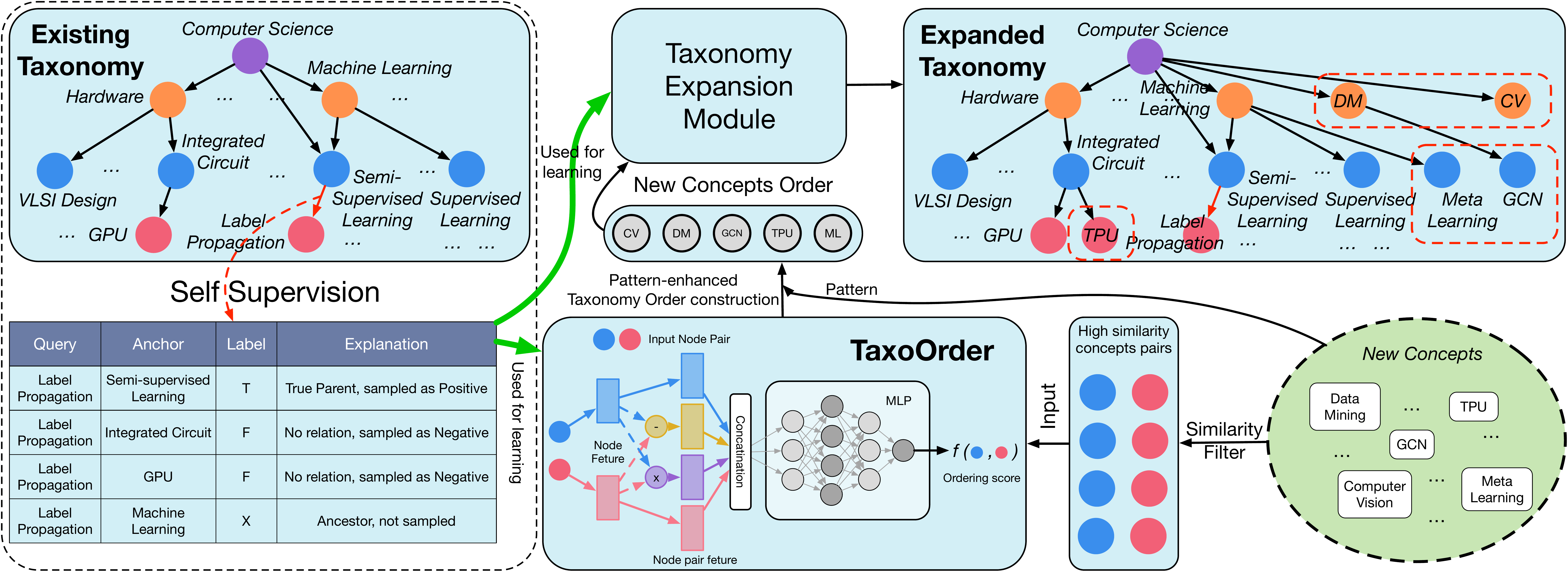}
    \caption{Overview of the \TaxoOrderBf framework. \TaxoOrderBf leverage self-supervision on existing taxonomy and use pattern-enhanced sorting algorithm to provide an order. Such order was then used by taxonomy expansion modules.}
    \label{fig:framework}
\end{figure*}
\section{The \TaxoOrderBf Framework}
In this section, we first give an introduction of our pattern-enhanced concept sorting model. Then we elaborate on the \TaxoOrder design and learning details. Finally, we explain how to expand the taxonomy with expansion models\footnote{Here we leverage two taxonomy expansion models, \Name{TaxoExpan} and \Name{Arborist}. Such expansion work as follows: for a given query concept, find the most likely parent node in the existing taxonomy and assign the query concept as the child node for such parent node.} based on the order of new concepts generated by \TaxoOrder. The overall illustration of the proposed \TaxoOrder framework is shown in Figure \ref{fig:framework}.

\subsection{Pattern-based Concept Graph}
For the new concept set $\mathcal{C}$, we first generate some hypernyms relations by patterns in surface names (e.g., ``science" is a hypernym of ``computer science", ``text mining" is a hypernym of ``biotext mining"). Such surface matching process provides a set of high-quality hypernym-hyponym edges ($\E_{pattern}$).
The generated edges, as well as the set of new concepts $\C$, lead to the concept graph $\mathcal{G}_{concept}=(\C, \E_{pattern})$. However, such graph contains minor noisy edges and potentially forms cycles within graph which prevents the graph to be sorted. Then to provide an order for each concept and to best preserve the high-quality order pairs mined by the rule-based surface matching, cycles are cut in decreasing order of cycle size to form $\mathcal{T}_{concept}$, which is guaranteed to be a DAG. 
Such pattern-based concept graph generation enjoys high precision but has low recall: the coverage of valid hypernym-hyponym pairs is far from satisfactory. To leverage the existing taxonomy, we then model the concept pairs and learn a concept pair sorting algorithm to further enhanced the order relation among new concepts.

\subsection{Modeling Concept Pairs}\label{sec:model concept pair and goal}
The concept sorting model we aim to learn is a scoring function $f$ (as shown in Figure \ref{fig:framework}), which inputs a pair of concepts $(a,c)$ and outputs their relative order score. 
Higher score $f(a,c)$ indicates higher confidence to have pair $(a , c)$ ordered as $a$ in front of $c$.
In other words, $a$ is a more general concept than $c$ and therefore should be inserted into taxonomy first.


Following the previous work \cite{Shen2020TaxoExpanST}, we assume each new concept has an initial feature vector learned from the associated corpus. 
Concept $c_i$ is represented using its initial feature vector $\mathbf{c_i}\in\mathbb{R}^{d}$.
For each candidate concept pair $\langle a,c \rangle$, we generate four sub-features: 
\begin{itemize}
    \item $\mathbf{a}$: Embedding of concept ``a'';
    \item $\mathbf{c}$: Embedding of concept ``c'';
    \item $\mathbf{a-c}$: Difference between embedding of concept ``a'' and concept ``c'';
    \item $\mathbf{a} \odot \mathbf{c}$: Element-wise multiplication of embedding of concept ``a'' and concept ``c''.
\end{itemize}
The feature vector $\textsc{Feature}(a,c)$ is the concatenation of four sub-features mentioned above:
\begin{equation}
    \small
    \textsc{Feature}(a,c) = [ \mathbf{a} \mathbin\Vert \mathbf{c} \mathbin\Vert \mathbf{a-c} \mathbin\Vert \mathbf{a}\odot \mathbf{c}].
\end{equation}
Then we parameterize the scoring function $f$ as a multi-layer perceptron (MLP), which inputs the feature vector $\textsc{Feature}(\cdot,\cdot)$ of a pair of concepts.
Section \ref{sec:MLP} describes the detailed implementation of the scoring function.

\subsection{Self-supervision Generation}
To learn the concept sorting model, we generate self-supervisions as shown in Figure \ref{fig:framework}. Given one edge $(n_p,n_c)$ in the existing taxonomy $\mathcal{T}$, we construct N negative pairs by fixing the child node $n_c$ and randomly selecting N concept nodes $n_1,n_2$ ... $n_N$ which are not ancestors of $n_c$. For example, as illustrated in the ``Self-Supervision'' part in Figure \ref{fig:framework}, for a query concept ``Label Propagation'', there is a real edge (Semi-Supervised Learning, Label Propagation) in the taxonomy. This real edge corresponds to a positive sample. Then to generate the negative samples, we fix ``Label Propagation'' and select ``Integrated Circuit'' and ``GPU'' as negative samples. Here, the node ``Machine Learning'' will not be sampled because it is the ancestor of ``Label Propagation''. These N+1 pairs collectively consist of one training instance $\boldsymbol{X} = \{(n_p,n_c),(n_1,n_c),(n_2,n_c),\dots,(n_N,n_c)\}$. We repeat the process above for each edge in the existing taxonomy $\mathcal{T}$ to generate the full self-supervision data $\mathbb{X} = \{ \boldsymbol{X}_1, \boldsymbol{X}_2, \dots, \boldsymbol{X}_{|\mathcal{E}|} \}$.

\subsection{Model Training}
In self-supervised learning settings, contrastive loss is widely adopted and Noise Contrastive Estimation (\Name{NCE}) \cite{Gutmann2010NoisecontrastiveEA} provides great discriminative learning power. Since in \TaxoOrder settings, the negative samples are readily available in a large number, we learn our \TaxoOrder on $\mathbb{X}$ using the \Name{InfoNCE} loss \cite{Oord2018RepresentationLW} as follows:
\begin{equation}
    \mathcal{L}(\Theta) = -\frac{1}{|\mathbb{X}|} \sum_{\boldsymbol{X_i}\in \mathbb{X}} \Bigg[log \frac{f(n_p,n_c)}{\sum_{(n_j,n_c) \in \boldsymbol{X_i}}f(n_j,n_c)}\Bigg],
    \label{eq:loss}
\end{equation}
where, the subscript $j \in [p,1,2,...,N]$. If $j=p$, $(n_j,n_c)$ is a positive pair, otherwise it is a negative pair. The above loss is the cross entropy of classifying positive pair $(n_p,n_c)$ correctly, with $\frac{f(n_p,n_c)}{\sum_{(n_j,n_c)\in \mathbb{X}_i} f(n_j,n_c)}$ as the model prediction.

\begin{algorithm}[htbp]
   \caption{Self-supervised learning of \TaxoOrder}
   \label{algo:self_train}
   \KwIn{
     A taxonomy $\T^{0}$; negative size $N$, batch size $B$; model $f(\cdot|\Theta)$.
   }
   \KwOut{Learned model parameters $\Theta$.}
   Randomly initialize $\Theta$\;
   \While{$\mathcal{L}(\Theta)$ in Eq. (\ref{eq:loss}) not converge} {
   	Enumerate nodes in $\T^{0}$ and sample $B$ nodes without replacement\;
	$\mathbb{D} = \emptyset$ \# current batch of training instances\;
	\For{each sampled node $n_q$} {
	    Select one of its parents $n_p$ to construct one positive pair $\pair{n_p, n_q}$.
	    
		Generate $N$ negative pairs $\{\pair{n_{p}^{l}, n_q}, \dots, \pair{n_{p}^{N}, n_q}\}$\;
		
		$\mathbb{D} \gets \mathbb{D} \cup \{\pair{n_p,n_q}, \pair{n_{p}^{1}, n_{q}}, \dots, \pair{n_{p}^{N}, n_{q}}\}$\;
	}
	Update $\Theta$ based on $\mathbb{D}$.
   }
   Return $\Theta$\;
 \end{algorithm}

\subsection{Candidate Concept Pair Generation}
\label{sec:score}
At the inference stage, we are given a set of new query concepts $\mathcal{C}$ and aim to apply the learned model $f (\cdot|\Theta)$ to predict the pair-wise ordering of concepts in $\mathcal{C}$. 
Notably, we only need to correctly sort relevant concepts. 
That is, we aim to place hypernym concept in the front of its hyponyms, while for irrelevant concept pairs, their ordering will not affect the resulting taxonomy.
Thus, we first examine the semantic similarity between each pair of new concepts and only consider the pairs with high semantic similarity scores as candidate concept pairs, because they are more likely to have hypernym-hyppnym relations. To achieve this goal, we perform straightforward threshold-based filtering to get high quality candidate pairs. Such threshold is learned from the existing taxonomy. The details about the threshold setting is illustrated in Section \ref{sec:thd}.

\begin{figure}[htbp]
    \centering
    \includegraphics[width=\linewidth]{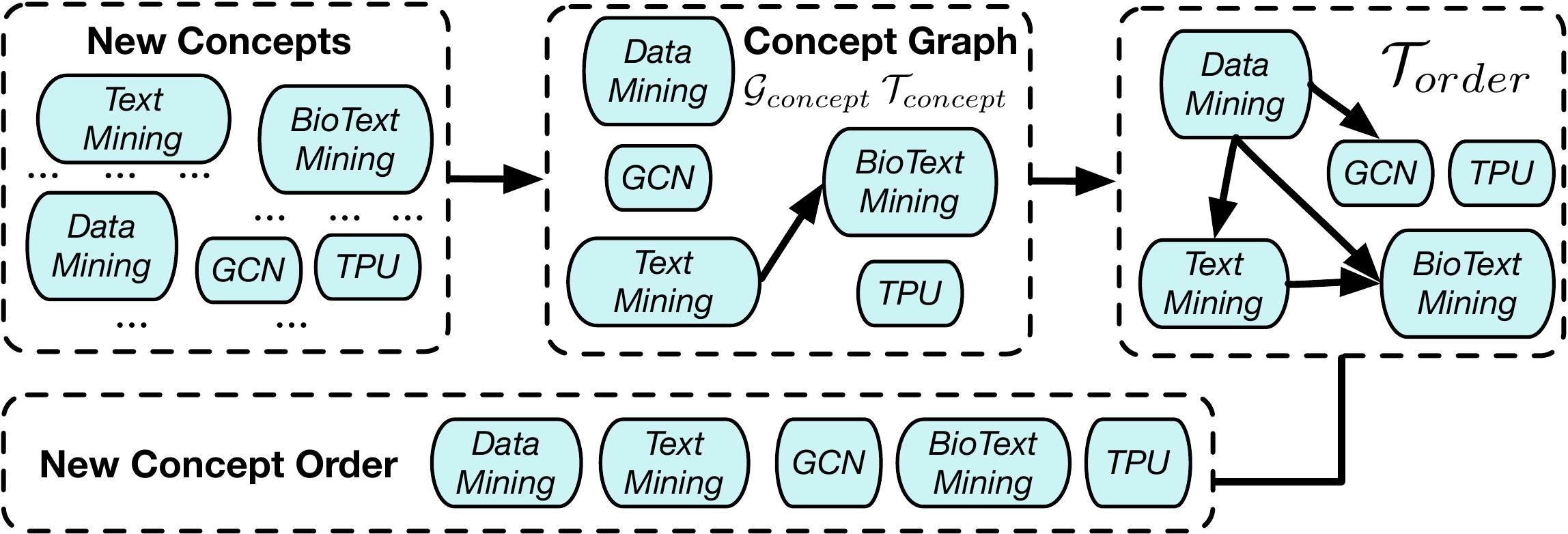}
    \caption{Illustration for pattern-enhanced concept sorting.}
    \label{fig:det}
\end{figure}

\subsection{Pattern-enhanced Concept Sorting}
Our \TaxoOrder learns a relative score for each concept pair which indicates the relative ranking order within the new concept set. 
Thus, we generate the pseudo-edges $\langle n_p, n_c\rangle$ with weight $f(n_p,n_c)$ to enforce the order with confidence learned in \TaxoOrder.

Finally, we generate the order leveraging both pattern-based concept graph $\mathcal{T}_{concept}$ and pseudo-edges $\langle n_p, n_c\rangle$.
To maintain the DAG property, pseudo-edges are iteratively appended  to $\mathcal{T}_{concept}$ in the descent order of their weights. 
During the $ith$ iteration, if pseudo-edge $e_i$ forms cycle in current $\mathcal{T}^{i}_{order}$, it will be discarded. After appending all pseudo-edges, the $\mathcal{T}_{concept}$ finally becomes $\mathcal{T}_{order}$. The order $O_{\mathcal{C}}$ is obtained by applying topological sort on $\mathcal{T}_{order}$.
\begin{equation}
\label{eq:sort}
    O_{\mathcal{C}} = \textsc{TopologicalSort}(\mathcal{T}_{order})
\end{equation}
That enforces the output order $O_{\mathcal{C}}$ follow the pair-wise ordering rule given by the edges in $\mathcal{T}_{order}$. An illustrative example for the whole pattern-enhanced concept sorting process is provided in Figure \ref{fig:det}.
\subsection{Iterative Taxonomy Expansion}
For expanding existing taxonomies, we adopt \TaxoExpan \cite{Shen2020TaxoExpanST} and \Arborist \cite{Manzoor2020ExpandingTW}, two state-of-the-art taxonomy expansion systems. 
We iteratively insert new concepts following the order $O_{\mathcal{C}}$ given by Eq.~\ref{eq:sort}. 
Specifically, at iteration $t$, we update the current taxonomy $\T^{t-1}=(\N^{t-1}, \E^{t-1})$ by attaching the $t$-th concept ($c^t$) to the concept $n^*\in\N^{t-1}$ with the highest matching score:
\begin{equation}
    \mathcal{T}^{t} = (\N^t=\N^{t-1}\cup\{n^*\}, \E^t=\E^{t-1} \cup (n^*, c^t)),
    \label{eq:T update}
\end{equation}
\begin{equation}
    n^* = \textsc{Expansion}(\mathcal{T}^{t-1},c^t)
    \label{eq:TaxoExpan module}
\end{equation}
where ($n^*$) is output by the expansion model,

\begin{table}[htbp]
    \centering
    \caption{Dataset Statistics. $|\mathcal{N}|$ and $|\mathcal{E}|$ are the number of nodes and edges in the existing taxonomy. $|\mathcal{D}|$ indicates the taxonomy depth and $|\mathcal{C}|$ is the number of new concepts.}
    \begin{tabular}{ ccccc }
    \toprule
    \textbf{Dataset} & {$|\mathcal{N}|$} & $|\mathcal{E}|$ & $|\mathcal{D}|$& $|\mathcal{C}|$\\
    \midrule
    \textbf{MAG-CS} & 24,754 & 42,329 & 6 & 3,765\\
    \textbf{MAG-Full} & 355,808 & 638,674 & 6 & 37,804\\
    \bottomrule
    \end{tabular}
    
    \label{tab:dataset}
\end{table}

\section{Experiments}

In this section, we discuss the information about the dataset and then make comparisons between different methods for our task.
\begin{figure}[htbp]
    \centering
    \includegraphics[width=\linewidth]{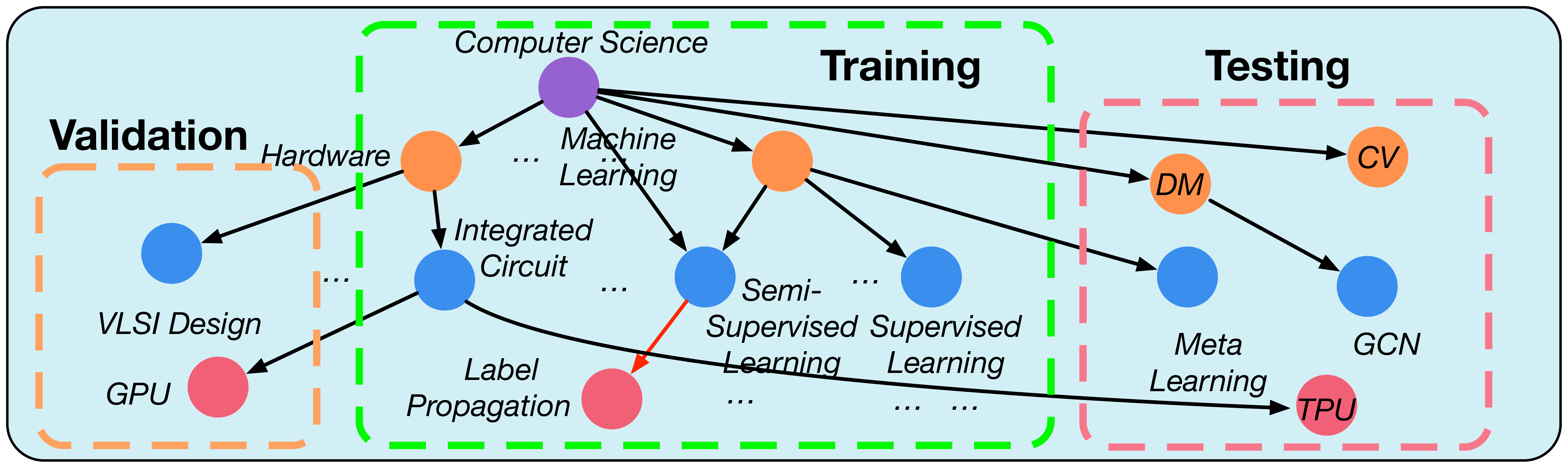}
    \caption{Partition scheme of training set, validation set, and testing set.}
    \label{fig:train val test}
\end{figure}

\subsection{Dataset}

We evaluate \TaxoOrder on the public Field-of Study (FoS) Taxonomy\footnote{https://docs.microsoft.com/en-us/academic-services/graph/reference-data-schema} in Microsoft Academic Graph (MAG) \cite{mag}. 
We modify the dataset from \TaxoExpan in separation to fit our task. 
As shown in Figure \ref{fig:train val test}, we only mask leaf nodes for validation. For testing, if some node concept $c$ is sampled, we mask the whole DAG rooted at $c$ to construct the testing set. Table \ref{tab:dataset} shows the statistics of the two datasets mentioned above.

In detail, the FoS taxonomy contains more than 660k scientific concepts and over 700k taxonomic relations. Although it is constructed in a  semi-automatic manner, the previous study \cite{shen-etal-2018-web} shows that this taxonomy is of high quality. We remove all concepts that have no relation in the original FoS taxonomy and then randomly mask 20\% of concepts (along with their relations) for validation and testing. The remaining FoS taxonomy is then treated as the input existing taxonomy. We refer to this dataset as \textbf{MAG-Full}. Based on MAG-Full, we construct another dataset called \textbf{MAG-CS} focusing on the computer science domain. Specifically, we first select a subgraph consisting of all descendants of the ``computer science" node and then mask around 10\% of concepts in this subgraph for testing and then mask another 10\% leaf concepts for validation.

As discussed in Section \ref{sec:model concept pair and goal}, each concept within our task has an initial feature vector. To obtain such feature vectors, we first construct a corpus that consists of all paper abstracts mentioning at least one concept in the original MAG dataset. Specifically, each concept should be treated as a single token for embedding learning (e.g., for multi-word expression ``data mining'', it will be converted to ``data\_mining''). 
Then, we learn 250-dimension word embedding as initial feature vectors using the skip-gram model word2vec \cite{mikolov2013distributed}. 
\begin{figure}
    \centering
    \includegraphics[width=0.6\linewidth]{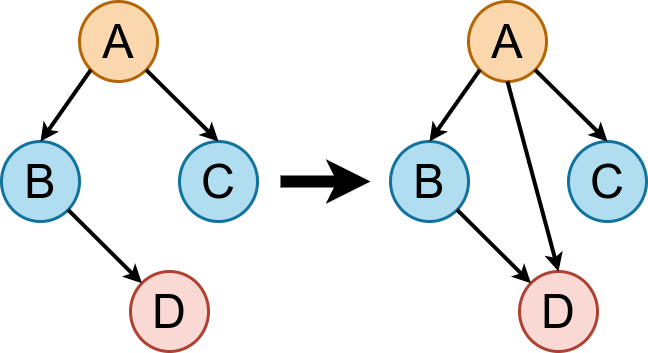}
    \caption{Ancestor Modification}
    \label{fig:ancestor}
\end{figure}

\subsection{Evaluation Metrics}
\label{sec:eval details}

\begin{itemize}
    \item \textbf{Error Node Count (ENC)} is the number of query concepts whose parent is not present in the existing taxonomy when it is inserted.
    \item \textbf{Hit@k} is the number of query concepts whose parent is ranked
in the top-$k$ positions, divided by the total number of queries.
    \item \textbf{Pred F1} is calculated based on the model predicted edges ($\mathcal{E}_{pred}$) and real edges ($\mathcal{E}_{gt}$). 
    $$\textbf{P}= \frac{|\mathcal{E}_{pred} \cap \mathcal{E}_{gt}  |}{|\mathcal{E}_{pred}|},\textbf{R}= \frac{|\mathcal{E}_{pred} \cap \mathcal{E}_{gt}  |}{|\mathcal{E}_{gt}|}$$
    \item \textbf{Edge F1} is calculated based on the expanded taxonomy ($\mathcal{T}_{pred}$) edges and ground truth taxonomy ($\mathcal{T}_{gt}$) edges. 
    $$\textbf{P}= \frac{|\mathcal{E}(\mathcal{T}_{pred}) \cap \mathcal{E}(\mathcal{T}_{gt})  |}{|\mathcal{E}(\mathcal{T}_{pred})|}, \textbf{R}= \frac{|\mathcal{E}(\mathcal{T}_{pred}) \cap \mathcal{E}(\mathcal{T}_{gt})  |}{|\mathcal{E}(\mathcal{T}_{gt})|}$$
    \item \textbf{Ancestor F1}: To better capture the quality of the taxonomy we modify both the expanded taxonomy ($\mathcal{T}_{pred}$) and ground truth taxonomy ($\mathcal{T}_{gt}$) by connecting each concept to its all ancestor concepts to form $\mathcal{T}^*_{pred}$ and $\mathcal{T}^*_{gt}$. This process is illustrated in Figure \ref{fig:ancestor}. 
    $$\textbf{P}= \frac{|\mathcal{E}(\mathcal{T}^*_{pred}) \cap \mathcal{E}(\mathcal{T}^*_{gt})  |}{|\mathcal{E}(\mathcal{T}^*_{pred})|}, \textbf{R}= \frac{|\mathcal{E}(\mathcal{T}^*_{pred}) \cap \mathcal{E}(\mathcal{T}^*_{gt})  |}{|\mathcal{E}(\mathcal{T}^*_{gt})|}$$
    \item \textbf{Mean Reciprocal Rank (MRR)} calculates the reciprocal rank
of a query concept’s true parent. We follow \cite{ying2018gcn} and use a scaled
version of MRR in the below equation:
\begin{equation}
    \small
    \textbf{MRR} = \frac{1}{|C|} \sum_{c \in C} \frac{1}{parent(c)} \sum_{i \in parent(c)} \frac{1}{\ceil*{R_{i,c}/10}}
\end{equation}
where $parent(c)$ represents the parent node set of the query concept $c$, and $R_{i,c}$ is the rank position of query concept $c$’s true
parent $i$. We scale the original MRR by a factor of 10 to
amplify the performance gap between different methods.
\end{itemize}

\subsection{Compared Methods}
We compare our \TaxoOrder concept sorting algorithm with several baselines. All these method are used to determine the inserting order for new concepts. We examine the effectiveness of these methods with two taxonomy expansion modules. Note that taxonomy expansion module is independent to our concept sorting problem, thus make our \TaxoOrder compatible with other original expansion work.
\begin{enumerate}
    \item \NameBf{Random}: This method simply inserts new concepts in random order which applies the \textsc{Expansion} model directly to the new task.
    \item \NameBf{Affinity}: The \textsc{Expansion} model assigns affinity scores for the candidate $\langle n_p, n_q \rangle$ pair. Such affinity scores can be interpreted as the level of confidence that $n_p \in \T$ is the hypernym of $n_q \in \C$. \NameBf{Affinity} inserts new concepts based on the affinity scores from the \textsc{Expansion} module. \NameBf{Affinity} sorts the new concepts by the highest affinity score available for each query $n_q$: the new concept node with the highest affinity score with the existing taxonomy node gets inserted first.
    \item \NameBf{MLP}: This is the proposed \TaxoOrderBf without pattern guidance, the DAG $\mathcal{T}_{order}$ is generated by applying pruning algorithm (Minimum Spanning Tree) on $\mathcal{G}_{order}$ .
    \item \NameBf{Pattern}: This method first performs \textsc{TopologicalSort} on $\mathcal{T}_{pattern}$ to determine the order $\O$. Then it inserts the new concepts using the topological order $\O$ of $\mathcal{T}_{pattern}$.
    \item \TaxoOrderBf: Proposed pattern-enhanced \TaxoOrder method.
\end{enumerate}

\begin{table*}[htbp]
    \centering
    \caption{Overall results on MAG-CS datasets with \NameBf{TaxoExpan} and \NameBf{Arborist}. Note that smaller ENC indicates better model performance. For all other metrics, larger values indicate
better performance. We highlight the best two models in terms of the performance under each metric.}
    \begin{tabular}[width=\linewidth]{ c|ccccccc }
    \toprule
    \textbf{Methods} & \textbf{ENC}  & \textbf{MRR}  & \textbf{Hit@1} & \textbf{Hit@3}  & \textbf{Pred F1} & \textbf{Edge F1} & \textbf{Ancestor F1}\\
    \midrule
    \Name{TaxoExpan}+\textbf{Ground Truth}& \textbf{0}     & \textbf{0.2702 }    & \textbf{0.1934 } & \textbf{0.2576 }   & \textbf{0.1427 } & \textbf{0.9026 } & \textbf{0.9345 }\\
    \midrule
    \Name{TaxoExpan}+\Name{Random}      & 1208  & 0.2113     & 0.1450  & 0.2109    & 0.1070  & 0.8985  & 0.9328 \\
    \Name{TaxoExpan}+\Name{Affinity}    & 966   & 0.2169     & 0.1498  & 0.2157    & 0.1105  & 0.8989  & 0.9329 \\
    \Name{TaxoExpan}+\Name{MLP}         & 794   & 0.2320     & 0.1610  & 0.2117    & 0.1188  & 0.8987  & 0.9326 \\
    \Name{TaxoExpan}+\Name{Pattern}     & 1892  & 0.1662     & 0.1116  & 0.1687    & 0.0823  & 0.8957  & 0.9327 \\
    \Name{TaxoExpan}+\TaxoOrderBf & \textbf{437}   & \textbf{0.2595}    & \textbf{0.1782} & \textbf{0.2534}   & \textbf{0.1315} & \textbf{0.9013} & \textbf{0.9332} \\
    \midrule
    \Name{Arborist}+\textbf{Ground Truth}& \textbf{0}     & \textbf{0.2272}    & \textbf{0.2024} & \textbf{0.2619}   & \textbf{0.1493} & \textbf{0.9034} & \textbf{0.9315}\\
    \midrule
    \Name{Arborist}+\Name{Random}      & 1208  & 0.1877    & 0.1450 & 0.2135   & 0.1143 & 0.8994 & 0.9305\\
    \Name{Arborist}+\Name{Affinity}    & 966   & 0.1919    & 0.1482 & 0.2183   & 0.1150 & 0.8995 & 0.9305\\
    \Name{Arborist}+\Name{MLP}         & 794   & 0.2035    & 0.1620 & 0.2316   & 0.1250 & 0.9006 & 0.9304\\
    \Name{Arborist}+\Name{Pattern}     & 1892  & 0.1514    & 0.1222 & 0.1740   & 0.0902 & 0.8966 & \textbf{0.9309}\\
    \Name{Arborist}+\TaxoOrderBf & \textbf{437}   & \textbf{0.2258}    & \textbf{0.1780} & \textbf{0.2584}   & \textbf{0.1313} & \textbf{0.9013} & 0.9300\\
    \bottomrule
    \end{tabular}
    \label{tab:result}
\end{table*}

\subsection{Implementation Details}
\subsubsection{Threshold for Semantic Similarity}
\label{sec:thd}
\begin{figure}[htbp]
    \centering
    \subfloat[real edges]{{\includegraphics[width=0.45\linewidth]{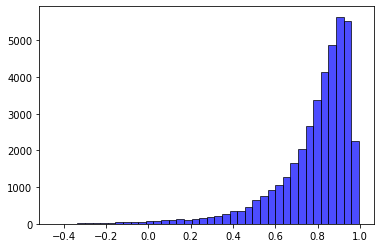} }}%
    \subfloat[random samples]{{\includegraphics[width=0.45\linewidth]{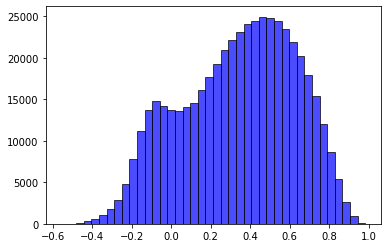} }}%
    \caption{embedding cosine similarity distribution between (a) real edges in existing taxonomy $\mathcal{T}^0$ and (b) random sampled node pair from $\mathcal{N}(\mathcal{T}^0)$}
    \label{fig:thd}
\end{figure}

As described in Section~\ref{sec:score}, we only consider concept pairs whose semantic similarity is higher then a threshold $\alpha$.
Here, we adopt cosine similarity, and the threshold $\alpha$ is set based on the existing taxonomy to avoid heavy tuning.
To set the threshold $\alpha$, we compare the cosine similarity distribution between (1) real edges in existing taxonomy $\mathcal{T}^0$ and (2) randomly sampled node pair from $\mathcal{N}^0$.
Figure \ref{fig:thd} shows that there is a clear difference between the distribution of similarity scores of real edges and that of randomly generated node pairs. 
Hence, we simply use the mean of the similarity scores of real edges 0.7 as our threshold $\alpha$. 

\subsubsection{MLP Details}
\label{sec:MLP}
We simply use one hidden layer MLP and the structure is described by formula \ref{eq:MLP} below,
\begin{equation}
    \small
    f^{\Name{MLP}} (a,c) = \sigma (\mathbf{W}_2\gamma (\mathbf{W}_1 \times feature(a,c) +\mathbf{B}_1) +\mathbf{B}_2)
    \label{eq:MLP}
\end{equation}
where $\mathbf{W}_1,\mathbf{W}_2,\mathbf{B}_1$ and $\mathbf{B}_2 $ are learnable parameters, $\sigma$ is a sigmoid function, and $\gamma$ is ReLU activation function.

Empirically, we use 512-dimensional hidden layer and 250-dimen\-sional embedding for node feature, which gives us $4 \times 250 \times 512 + 512\times1 + 512 + 1 = 513025$ parameters. We even further studied a deeper MLP with 2.5M parameters. They end up with similar performances. Hence we do not go further here.
\begin{table}[htbp]
    \centering
    \caption{Average run-time analysis (hr/epoch)}
    \begin{tabular}{ccc}
    \toprule
         \TaxoOrder&\TaxoExpan&\Arborist  \\
    \midrule
         0.15&0.25&0.1\\
    \bottomrule
    \end{tabular}

    \label{tab:runtime}
\end{table}
\subsubsection{Code Reproducibility and Runtime Analysis}
For taxonomy expansion modules, we modified \TaxoExpan\footnote{https://github.com/mickeystroller/TaxoExpan} and \Arborist\footnote{https://github.com/cmuarborist/cmuarborist-core} to fit the new task. For the \TaxoOrder module, we use PyTorch \cite{paszke2019pytorch} as the basic framework. For \TaxoExpan and \TaxoOrder, we train, validate, and test the model on a Linux Host with Intel(R) Core(TM) i7-8700K CPU @ 3.70GHz CPU, 48G DDR4 Memory, and Nvidia GeForce RTX 2080 GPU. For the \Arborist module, we run on a Linux Server with Intel(R) Xeon(R) CPU E5-2680 v2 @ 2.80GHz CPU and 256G Memory. The average run-time per epoch for each sub-module is listed in Table \ref{tab:runtime}.

\subsection{Experimental Results}
\smallskip
\noindent \textbf{Overall Performance.}
Table \ref{tab:result} shows the overall result of all compared methods. First, by comparing the correlation between ENC and other metrics, it can be concluded that the taxonomy expansion performance is highly sensitive to the inserting order: the lower the ENC, the higher the expansion performance can achieve. And the ground truth inserting order gives the best expansion performance since its ENC is zero. 
Among all the methods compared in both expansion modules in experiments, the \TaxoOrder outperforms the baseline methods and is distinguished out by a large margin. 
Even compared with the ground truth order, the \TaxoOrder performs relatively well in terms of most of the evaluation metrics. 
Comparing \Name{Affinity} with \Name{Random}, the ENC has dropped a lot, but in the expansion quality evaluation, there is only a small improvement from \Name{Random} to \Name{Affinity}. That is because the \Name{Affinity} score is not tailored to this sorting task. Although it represents the hypernym-hyponym relation to some extent, the lack of learning process on this sorting task limits its capability to get better expansion results.

\smallskip
\noindent \textbf{Ablation Study.}
We perform the ablation study by comparing \TaxoOrder with \Name{MLP} and \Name{Pattern} respectively. 
For the \Name{MLP} model, the ENC is significantly smaller than randomly assigned order, which demonstrates the feasibility of the proposed learning framework, however, the \Name{MLP} model produces extra noisy data which also violates the real edge in the new concepts. 
These noisy data make the inserting result worse than \TaxoOrder.
By comparing between \TaxoOrder and \Name{Pattern}, the Pattern method provides many high-quality potential hypernyms relations but not all parent-children concepts have this surface name matching rule. That leaves many concepts orphan in the DAG created in \Name{Pattern}. The \textsc{TopologicalSort} has no control over the order of these orphan concepts, thus making the ENC of the \Name{Pattern} even worse than \Name{Random}. Other evaluation metrics follow the same logic, higher ENC gives worse expansion results. But the \Name{Pattern} model does contain high-quality order information for the concepts with matching results. \TaxoOrder model makes a big step and benefits both from the pattern-based method for high-quality order extraction and also covers most of the new concepts such that decrease the ENC and achieve nearly comparable with ground truth order.

\section{Related Work}

\smallskip
\noindent \textbf{Taxonomy Construction.}
Traditional taxonomy construction me\-thods use lexical features from the resource corpus such as lexical-patterns \cite{Nakashole2012PATTYAT,jiang2017metapad,hearst-1992-automatic,agichtein-2000-snowball} or embedding-based distribution methods \cite{jin2018junction,luu-etal-2016-learning,roller-etal-2014-inclusive,weeds-etal-2004-characterising}. Later work CRIM \cite{bernier-colborne-barriere-2018-crim} utilized word-embedding, negative sampling, fine-tuning, and multiple projection matrices to achieve the best performance in the SemEval 2018 hypernym discovery task.

\smallskip
\noindent \textbf{Taxonomy Expansion.}
\Arborist \cite{Manzoor2020ExpandingTW} follows the framework in piecewise projection-learning on word-embeddings \cite{fu-etal-2014-learning} and designed a training objective with a marginal loss which enforces the projection matrix learning the hypernym-hyponym relationship \cite{camacho-collados-etal-2018-semeval}. 
With graph neural networks being introduced, \TaxoExpan \cite{Shen2020TaxoExpanST} utilizes a position-enhanced graph neural network (GNN) that captures the local structure of a concept node in the existing taxonomy and at the same time trains a matching module to find the matching score for each anchor-query concepts pair measuring the confidence that the anchor is the hypernym of this query. 
And they claim a noise-robust training objective that enables the learned model to be insensitive to the label noise in the self-supervision data. 
\Name{STEAM} \cite{Yu2020STEAMST} further explores on top of the GNN based method and proposes to utilize semantic mini-paths in the existing taxonomy to further capture the hypernym-hyponym relations. 
\Name{TMN} \cite{Zhang2021TaxonomyCV} investigates such a problem in an alternative way, instead of one-to-one matching in the existing taxonomy expansion work, it proposes one-to-pair matching and introduces a channel-wise gating function to capture the hypernym and hyponym of query concepts.

\smallskip
\noindent \textbf{Lexical Memorization.}
The lexical memorization does matter in the taxonomy expansion task. Some “super-hypernym” nodes may absorb many new concepts as children. 
Although \TaxoOrder is learned on the existing taxonomy, it is applied to new concepts only. 
It won’t get in touch with the existing taxonomy for data leakage. 
It can be argued that the learned model somehow memorized the existing taxonomy structure. 
It also learned how relative generality is for a given concept pair. 
Memorizing the existing taxonomy structure won’t directly affect the prediction of the order in new concepts. 
The experiments also support this argument; the model didn't learn a perfect expansion model. 
Instead, it only learned a relative generality score function. And the output order shows that this order benefits the expansion model a lot.

\section{Conclusions}
This work extends the taxonomy expansion task from discovering the hypernym-hyponym relations between existing concepts and new concepts to a more general form: discovering the additional relation within new concepts. To solve this extended task, a novel sorting framework is proposed that leverages self-supervision from the existing taxonomy and learns a pattern-enhanced taxonomy ordering model \TaxoOrder which helps capture the hypernym relations within new concepts. Combined with taxonomy expansion modules, \TaxoOrder is able to provide high-quality inserting order and discover the hypernym-hyponym relations within new concepts. The experiment results and the ablation study showed the overall superiority of the proposed method and the effectiveness of each sub-module in the \TaxoOrder model design. Interesting future work may include using the ordering function with expansion models to clean the existing taxonomy.

\bibliographystyle{ACM-Reference-Format}
\bibliography{ref}


\end{document}